# ENHANCING TARGETED TRANSFERABILITY VIA FEATURE SPACE FINE-TUNING

*Hui Zeng*[1,2], *Biwei Chen*[3], *and Anjie Peng*[1,2*]

[1]Southwest University of Science and Technology      [3]Beijing Normal University
[2]Guangdong Provincial Key Laboratory of Information Security Technology

## ABSTRACT

Adversarial examples (AEs) have been extensively studied due to their potential for privacy protection and inspiring robust neural networks. Yet, making a targeted AE transferable across unknown models remains challenging. In this paper, to alleviate the overfitting dilemma common in an AE crafted by existing simple iterative attacks, we propose fine-tuning it in the feature space. Specifically, starting with an AE generated by a baseline attack, we encourage the features conducive to the target class and discourage the features to the original class in a middle layer of the source model. Extensive experiments demonstrate that only a few iterations of fine-tuning can boost existing attacks' targeted transferability nontrivially and universally. Our results also verify that the simple iterative attacks can yield comparable or even better transferability than the resource-intensive methods, which rest on training target-specific classifiers or generators with additional data. The code is available at: *github.com/zengh5/TA_feature_FT.*

## 1. INTRODUCTION

Exploring adversarial examples (AEs) is helpful in demystifying deep neural networks (DNN) [1], identifying the DNNs' vulnerability [2], and protecting privacy [3]. The attacking ability of an AE is usually measured by its transferability, i.e., the chance to fool unseen models. Recently, numerous transferable attacks have emerged, e.g., advanced algorithms have been adopted to stabilize the optimization direction and prevent AEs from falling into poor local maxima [4–6].

While existing studies have pushed the boundary of transferability under the untargeted mode, few focus on targeted attacks. Targeted transferability is much more daunting since it requires unknown models outputting a specific label [7]. Tailored schemes for improving the transferability of targeted attacks have been proposed to fill the gap. For instance, resource-intensive attacks seek extra, target-specific classifiers [8] or generators [9] to optimize adversarial perturbations. However, when the number of targeted classes is enormous, the training time will be problematic. In contrast, other researchers find that integrating novel loss functions with conventional simple iterative attacks can also enhance targeted transferability [10–12]. Our work relates to this latter literature.

The wide gulf between the success rates of white-box/black-box targeted attacks suggests that simple iterative attacks heavily overfit the source model, a dilemma not unique to just targeted attacks. In the context of untargeted attacks, apart from the well-known data or model augmentation strategies [13–17], researchers recently resorted to feature space disruption to mitigate the overfitting issue [18–22]. Because early layers (closer to the input) are more class-specific and later layers are more model-specific, targeting the middle layers may alleviate the overfitting issue and thus enhance transferability.

Our preliminary experiments, however, show that directly extending feature space attacks to the targeted mode achieves litter since representing the target class as a single point in an internal layer is difficult. Frustrating as it is, this idea is helpful for targeted attacks. In this work, we find that fine-tuning an existing AE in the feature space can effectively enhance its targeted transferability. Precisely, starting with an AE generated by a baseline attack, we fine-tune it for a few iterations to encourage the target class-associated features and suppress the original class-associated features in an internal layer of the source model. We incorporate the proposed fine-tuning strategy with various state-of-the-art simple iterative attacks. Experiments on ImageNet demonstrate that the targeted transferability of all the considered attacks can be improved notably via a few iterations of feature space fine-tuning.

## 2. RELATED WORK

### 2.1. Transferable untargeted attack

An untargeted attack aims to mislead a DNN-based classifier $f()$ into making a wrong output, i.e., $f(I') \neq f(I)$ where $I$ is the original image and $I'$ is the adversarial one. As a common baseline, the iterative fast gradient sign method (IFGSM) [23] can be formulated as follows:

$$I'_{n+1} = Clip_{I,\epsilon}\{I'_n + \alpha sign(\nabla_{I'_n}J(I'_n, y_o))\} \quad (1)$$

where $I'_0 = I$, $\nabla_{I'_n}J()$ denotes the gradient of the loss function $J()$ with respect to $I'_n$, $y_o$ is the original label, and $\epsilon$ is the perturbation budget. Researchers have proposed a variety of improved algorithms for IFGSM, e.g., the momentum iterative method (MI) [4] integrates a momentum term into the iterative process. Diverse inputs method (DI) [13] and translation-invariant method (TI) [14] leverage data augmentation to prevent attack from overfitting a specific source model. Moreover, these enhanced schemes can be integrated for better transferability, e.g., Translation Invariant Momentum Diverse Inputs IFGSM (TMDI).

*This work was supported by the network emergency management research special topic (no. WLYJGL2023ZD003), the Opening Project of Guangdong Province Key Laboratory of Information Security Technology (no. 2020B1212060078), Sichuan Science and Technology Program (no. 2022YFG0321).

In addition to crafting AEs on the output layer, recent works have started to perturb the internal layers. [24, 25] try to maximize the distance between AE and the corresponding benign image on the feature space. [20] proposes to disrupt internal features according to their importance.

$$arg\min_{I'}\sum(\bar{\Delta}_k^l \cdot f_k(I')), \ \ s.t.\ ||I' - I||_\infty \leq \epsilon \qquad (2)$$

where $f_k()$ denotes the feature maps from the $k$-th layer, and $\bar{\Delta}_k^l$ is the aggregate gradient used to measure the importance of the features. [21] improves [20] by using the integrated gradient [26] in measuring the feature importance, and [22] improves [20] by adopting a patch-wise mask in calculating aggregate gradients. Unlike the output-level attacks, extending feature-level attacks to the targeted mode is not trivial because representing the target class as a single point in an internal layer is difficult.

## 2.2. Transferable targeted attack

A targeted attack misguides a classification model to a specific label $y_t$ as whatever the attacker intends, i.e., $f(I') = y_t$. Beyond a simple extension of its untargeted counterpart, at least two additional challenges are unique to targeted attacks. The first is gradient vanishing [10, 11], i.e., the gradient decreases as the attack progresses. The other one is the restoring effect of $y_o$ and other high-confidence labels when an AE is transferred to an unseen model [10, 12]. Hence, tailored considerations are necessary for transferable targeted attacks.

[10] replaces traditional cross-entropy (CE) loss with the Poincare distance loss to address the decreasing gradient problem and introduces a triplet loss to push the attacked image away from $y_o$. [11] uses the Logit loss in the attack and reports better transferability than the CE loss.

$$L_{Logit} = -l_t(I') \qquad (3)$$

where $l_t(\cdot)$ denotes the logit output with respect to $y_t$. Another non-trivial contribution of [11] is the finding that targeted attacks need significantly more iterations to converge than untargeted ones do. [12] argues that not only the original label $y_o$, but also other high-confidence labels should be suppressed for better transferability. Such an idea can be realized by updating AEs according to the following direction:

$$\nabla(l_t(I') - \beta_1 l_o(I')) - \beta_2 \nabla(\sum_{i=0}^{N_h} l_{high-conf,i}(I')) \perp \quad (4)$$

Here, the first term is used to enhance the confidence of $y_t$ and suppress $y_o$ simultaneously, and the second term is used to suppress other high-confidence labels, where '$\perp$' denotes only the orthogonal component (to the first term) is kept.

Compared with the simple transferable attacks reviewed above, resource-intensive attacks require training auxiliary target-class-specific classifiers or generators on additional data for better transferability. In the feature distribution attack [8], a light-weight, one-versus-all classifier is trained for each target class $y_t$ at each specific layer to predict the probability that a feature map is from $y_t$. [9] trains an input-adaptive generator to synthesize targeted perturbation and achieves state-of-the-art transferability. However, a dedicated generator must be learned for every (*source model*, *target class*) pair in [9]. Henceforth, we denote the traditional CE loss-based attack, the Po+Trip loss-based attack [10], the logit loss-based attack [11], the high-confidence suppressing attack [12], and the transferable targeted perturbation [9] as CE, Po+Trip, Logit, SupHigh, and TTP, respectively.

## 3. PROPOSED SCHEME

### 3.1. Motivation

It is well-accepted that attacking the feature space can alleviate an AE's overfitting issue (to the source model), thus improving its transferability across models. In an untargeted scenario, feature space attacks can be realized by pushing AE away from the clean image in the feature space. As the saying goes, *it's easier to pull down than to build up*. In the targeted mode, however, there is no single point which perfectly represents $y_t$ in the feature space that we can pull closer to. Though this challenge is partially addressed in [8] by training a one-versus-all auxiliary classifier for each $y_t$, such a strategy is somewhat impractical in situations where the number of classes is enormous. *Can we elevate the targeted transferability of simple iterative attacks via feature-space modification without model training on additional data?*

We believe that AEs crafted by simple iterative attacks already have reasonably strong targeted attack abilities, which their nearly perfect white-box success rates can partially verify. Their transferability is expected to improve if we can alleviate the overfitting issue. To this end, we try to fine-tune a crafted AE in the feature space. Beginning with an AE (Fig. 1(a)) generated by the CE attack with TMDI augmentation ($y_t = $ '*grey owl*', $\epsilon = 16$), we perturb it in the final layer of the third block of the source model ResNet50 (Res50) [27] according to a targeted version of (2).

$$\underset{I'_{ft}}{argmax}\sum(\bar{\Delta}_k^{l'} \cdot f_k(I'_{ft})), \ \ s.t.\ ||I'_{ft} - I||_\infty \leq \epsilon \qquad (5)$$

Fig. 1(b) draws $I'_{ft}$'s confidence of '*grey owl*' as functions of iteration number. Despite that the curves of different target models vary a lot, they all rise sharply towards higher confidence in the early stage, which means a few iterations of fine-tuning may help alleviate the overfitting issue of AEs crafted by the baseline attack. Proceed from this delightful observation, in the next section, we propose fine-tuning a simple iterative attack-generated AE in the feature space to improve its targeted transferability.

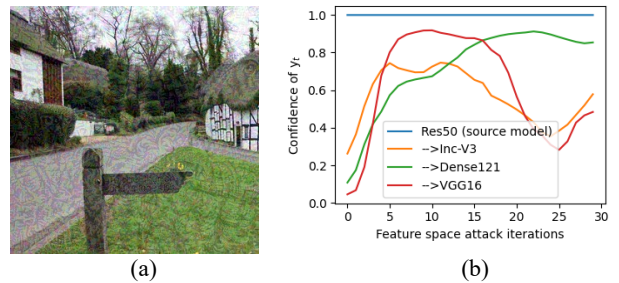

(a)                              (b)

**Fig. 1.** The overfitting issue is alleviated after feature space fine-tuninig. The target class is '*grey owl*' and the source model is Res50. (a) The AE, (b) the trend of the confidence with respect to '*grey owl*' as fine-tuninig progress. The step size is one.

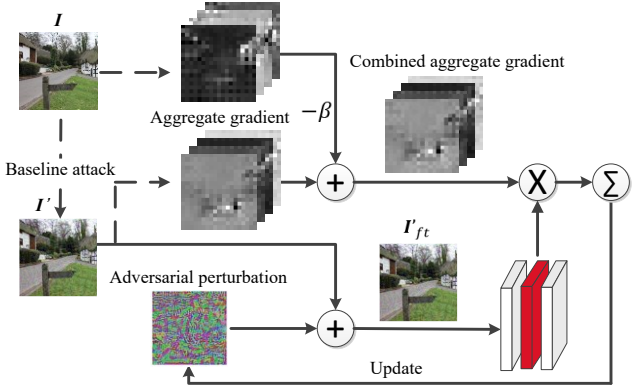

**Fig. 2.** Overview of feature space fine-tuning. Given an AE crafted by a baseline attack, its feature maps extracted from an internal layer (red block) are further optimized according to Eq. (7).

### 3.2. Feature space fine-tuning

Displayed in Fig. 2, Our proposed method starts from a benign image $I$ and an AE $I'$ generated by a baseline attack, e.g., CE or Logit, for $N$ iterations. Then, it calculates the aggregate gradient $\bar{\Delta}_k^{I',t}$ to $y_t$, from the $k$-th layer of the source model. Following the idea of suppressing the original label $y_o$ [10, 12], it also computes the aggregate gradient $\bar{\Delta}_k^{I,o}$, with respect to $y_o$, from $I$, and combines $\bar{\Delta}_k^{I',t}$ and $\bar{\Delta}_k^{I,o}$ to guide the subsequent feature space fine-tuning.

---

**Algorithm 1** Feature space fine-tuning

**Input**: A benign image $I$ with original label $y_o$; target label $y_t$.

**Parameter**: Iterations $N$ for baseline attack, $N_{ft}$ for fine-tuning.

**Output**: Adversarial image $I'_{ft}$.

1. Mount a baseline attack for $N$ iterations, and obtain an AE $I'$.
2. Calculate aggregate gradient $\bar{\Delta}_k^{I',t}$ from $I'$, and $\bar{\Delta}_k^{I,o}$ from $I$.
3. Obtain the combined aggregate gradient $\bar{\Delta}_{k,combine}$ as (6).
4. Fine-tune $I'$ as (7), for $N_{ft}$ iterations and obtain $I'_{ft}$.

$$\bar{\Delta}_{k,combine} = \bar{\Delta}_k^{I',t} - \beta \bar{\Delta}_k^{I,o} \qquad (6)$$

where $\beta$ is a predefined weight used to balance these two terms. We set $\beta = 0.2$, as in [12]. Finally, fine-tune $I'$ for $N_{ft}$ iterations, $N_{ft} \ll N$, with the optimization objective:

$$\underset{I'_{ft}}{argmax} \sum (\bar{\Delta}_{k,combine} \cdot f(I'_{ft})), \ s.t. ||I'_{ft} - I||_\infty \le \epsilon \quad (7)$$

In this manner, the features contributing to $y_t$ are encouraged, and those to $y_o$ are suppressed. **Algorithm 1** summarizes our procedure.

## 4. EXPERIMENTAL RESULTS

We contrast the proposed method with five simple iterative attacks: CE [23], Po+Trip [10], Logit [11], SupHigh [12], SU [28], and the current state-of-the-art generative attack, TTP [9]. All the iterative schemes start with the TMDI attack.

### 4.1. Experimental Settings

**Dataset.** Our experiments are conducted on the ImageNet-compatible dataset comprised of 1000 images [29]. All these images are cropped to $299 \times 299$ pixels before use.

**Networks.** Since transferring across different architectures is more demanding, we choose four pretrained models of diverse architectures: Inceptionv3 (Inc-v3) [30], Res50, DenseNet 121 (Dense121) [31], and VGG16bn (VGG16) [32] as in [11] to evaluate AEs' transferability.

---

**Table 1.** Targeted transfer success rate (%) without/with fine-tuning, in the single-model, random-target scenario. Best results are in **bold**.

| Attack | Source Model: Res50 | | | Source Model: Dense121 | | | Source Model: VGG16 | | | Source Model: Inc-v3 | | |
|---|---|---|---|---|---|---|---|---|---|---|---|---|
| | →Inc-v3 | →Dense121 | →VGG16 | →Inc-v3 | →Res50 | →VGG16 | →Inc-v3 | →Res50 | →Dense121 | →Res5 | →Dense121 | →VGG16 |
| CE | 3.9/**9.0** | 44.9/**60.4** | 30.5/**49.3** | 2.3/**13.2** | 19.0/**45.3** | 11.3/**34.8** | 0.0/0.0 | 0.3/**2.8** | 0.5/**3.3** | 1.8/**4.7** | 2.1/**7.8** | 1.5/**4.0** |
| Po+Trip | 7.1/**12.9** | 57.5/**68.1** | 36.3/**50.8** | 2.5/**8.9** | 15.2/**37.6** | 9.2/**29.7** | 0.1/**0.2** | 0.6/**2.9** | 0.6/**4.3** | 1.7/**5.0** | 3.3/**9.4** | 1.6/**4.9** |
| Logit | 9.1/**15.8** | 70.0/**75.3** | 61.9/**64.1** | 7.8/**15.1** | 42.6/**56.7** | 37.1/**49.3** | 0.8/**1.1** | 10.2/**15.5** | 13.6/**15.1** | 2.4/**6.3** | 5.0/**10.2** | 2.2/**7.7** |
| SupHigh | 9.6/**16.3** | 74.9/**75.7** | 63.5/**69.8** | 8.7/**15.9** | 48.1/**61.6** | 40.5/**52.2** | 0.8/**2.7** | 11.2/**16.2** | 13.6/**19.9** | 2.3/**5.9** | 4.5/**10.3** | 2.2/**9.6** |
| SU | 11.1/**18.6** | 72.5/**73.3** | 64.7/**68.5** | 10.0/**17.1** | 49.2/**62.3** | 43.1/**56.2** | 0.9/**1.7** | 13.7/**18.4** | 15.7/**20.3** | 3.0/**6.3** | 4.6/**11.2** | 3.5/**6.7** |

**Table 2.** Targeted transfer success rate (%) without/with fine-tuning, in the single-model, most difficult-target scenario.

| Attack | Source Model: Res50 | | | Source Model: Dense121 | | | Source Model: VGG16 | | | Source Model: Inc-v3 | | |
|---|---|---|---|---|---|---|---|---|---|---|---|---|
| | →Inc-v3 | →Dense121 | →VGG16 | →Inc-v3 | →Res50 | →VGG16 | →Inc-v3 | →Res50 | →Dense121 | →Res50 | →Dense121 | →VGG16 |
| CE | 1.3/**3.1** | 25.8/**45.3** | 15.0/**29.7** | 1.2/**6.1** | 6.5/**23.4** | 3.6/**19.2** | 0.0/0.0 | 0.0/**1.4** | 0.0/**0.6** | 2.4/**5.9** | 4.2/**7.8** | 2.3/**5.0** |
| Po+Trip | 2.8/**7.3** | 40.5/**50.6** | 20.5/**37.8** | 1.7/**4.2** | 6.1/**21.8** | 2.5/**17.7** | 0.0/**0.2** | 0.1/**1.2** | 0.0/**0.6** | 2.4/**6.1** | 4.1/**7.5** | 2.7/**6.3** |
| Logit | 3.6/**7.5** | 51.6/**53.1** | 38.6/**44.3** | 3.5/**8.3** | 22.7/**41.6** | 18.3/**37.5** | 0.3/**0.4** | 2.8/**8.9** | 7.0/**8.5** | 3.8/**8.0** | 5.5/**10.5** | 3.2/**7.9** |
| SupHigh | 4.0/**8.1** | 54.5/**57.9** | 41.6/**51.2** | 4.0/**8.6** | 24.5/**43.3** | 21.2/**40.4** | 0.1/**0.3** | 3.9/**10.1** | 6.8/**9.4** | 4.0/**8.2** | 4.9/**10.6** | 3.4/**7.2** |
| SU | 5.0/**7.5** | **56.2**/54.8 | 44.1/**49.7** | 4.4/**8.0** | 27.4/**41.7** | 24.3/**37.6** | 0.1/**0.2** | 5.7/**7.6** | 7.4/**9.7** | 3.9/**8.1** | 6.7/**12.1** | 3.9/**9.6** |

**Parameters.** For all competitors, the perturbations are restricted by $L_\infty$ norm with $\epsilon = 16$, and the step size is set to 2. Fine-tuning inevitably increases computational complexity even for $N_{ft} \ll N$. To make the running time comparable, we set the total iteration number $N=200$ for the baseline attacks without fine-tuning, and set $N=160$, $N_{ft}=10$ when fine-tuning is enabled. For the fine-tuning layer $k$, we opt for the middle layer, following [20, 21]. Specifically, we select to attack *Mixed_6b* for Inc-v3, the last layer of the third block for Res50 and Dense121, and *Conv4_3* for VGG16. Due to page limitation, we detail the ablation study on $N_{ft}$, $k$, and aggregate gradient method, and the result of data-free targeted UAP [11] in the supplementary file: *TA_feature_FT/supp.pdf*.

## 4.2. Single-model transfer

Table 1 reports the targeted transferability (random-target) across different models. All the baseline attacks benefit from the proposed feature space fine-tuning. The improvement is particularly salient for the CE attack. For example, when transferring from Dense121 to VGG16, the success rate almost triples after fine-tuning. Another major improvement is the case of Inc-v3, which was previously reported in [11, 12] to be challenging to transfer from or to. The last column confirms that fine-tuning at least doubles the success rates when Inc-v3 is the source model.

As suggested in [11], we consider a worst-case transfer scenario in which the target label is the least likely one. Table 2 compares different attacks: the improvement from fine-tuning is even more remarkable than the random-target scenario. Consistent with [12], we find that the success rates of the most difficult-target scenario are not necessarily lower than those of the random-target scenario when Inc-v3 is the source model before or after fine-tuning.

## 4.3. Ensemble transfer

Next, we evaluate the proposed fine-tuning scheme in the ensemble-model scenario. Specifically, we take turns taking out a model as the target and crafting AEs on the ensemble of the remaining models with equal weights. Note that there is no architectural overlap between the source and target models. Table 3 presents the targeted transferability in this scenario. Even though AE's transferability in the ensemble-model scenario has been significantly improved compared with that in the single-model scenario, the proposed fine-tuning scheme is still helpful for it.

## 4.4. Iterative vs. generative attacks

**Table 3.** Targeted transfer success rate (%) without/with fine-tuning in the ensemble-model scenario, where '−' indicates the hold-out model.

| Attack | −Inc-v3 | −Res50 | −Dense121 | −VGG16 | Average |
|---|---|---|---|---|---|
| CE | 24.4/**40.1** | 53.5/**57.6** | 77.3/**78.7** | 76.8/**79.3** | 58.0/**63.9** |
| Po+Trip | 22.5/**39.4** | 43.7/**48.9** | 71.9/**74.2** | 64.3/**68.7** | 50.6/**57.8** |
| Logit | 30.7/**41.4** | **68.8**/65.5 | **79.0**/76.9 | 81.6/**82.1** | 65.0/**66.5** |
| SupHigh | 34.8/**43.5** | **72.4**/70.5 | **81.8**/80.3 | **82.7**/82.2 | 67.9/**69.1** |
| SU | 35.8/**40.9** | **73.2**/70.8 | **82.1**/81.5 | **82.9**/81.4 | 68.5/**68.7** |

**Table 4.** Targeted transfer success rate (%) ($\epsilon = 8/16$) of fine-tuned iterative attacks vs. TTP, averaged over 10 targets. Source model: Res50.

| Attack | →Inc-v3 | →Dense121 | →VGG16 | Average |
|---|---|---|---|---|
| CE+*ft* | 1.9/14.5 | 33.1/63.5 | 27.8/51.3 | 20.9/43.1 |
| Po+Trip+*ft* | 3.1/17.5 | 39.2/68.7 | 33.4/57.6 | 25.2/47.9 |
| Logit+*ft* | 2.7/18.6 | 41.3/77.9 | 38.5/72.7 | 27.5/56.4 |
| SupHigh+*ft* | 3.7/19.9 | **44.1**/78.3 | **44.4**/76.9 | **30.7**/58.4 |
| TTP | **5.7**/**39.8** | 38.6/**79.5** | 44.2/75.4 | 29.5/**64.9** |

Last, we compete the fine-tuned simple iterative attacks against the state-of-the-art generative attack, TTP [9]. TTP asks for training a generator for each target label and each source model, which means $4 \times 1000$ generators are required to perform the random or most difficult-target attack as done in Section 4.2. Alternatively, we download ten pre-trained generators (Res50 being the discriminator during training) and follow the '*10-Targets (all source)*' setting of [9], which corresponds to simple iterative attacks targeting ten selected classes with Res50 as the source model. We evaluate all competitors under different $\epsilon$s. As shown in Table 4, fine-tuned state-of-the-art iterative attacks, Logit+*ft* and SubHigh+*ft*, yield comparable ($\epsilon=16$) or even better ($\epsilon=8$) transferability than TTP. As pointed out in [11], the relatively poor transferability of TTP under a low budget is because it heavily hinges on semantic patterns, which can be observed from Fig. 3(d). In contrast, the perturbation introduced by iterative methods resembles noise (Fig. 3(b, c)), which is less suspicious under human inspection. More examples are provided in the supplementary file.

## 5. CONCLUSION

We propose fine-tuning a given AE in the feature space to uplift its targeted transferability. The proposed scheme delicately combines the idea of feature-level perturbation with simple iterative attacks, effectively alleviating the overfitting issue in existing methods without training target-specific classifiers or generators. The superiority of the proposed fine-tuning scheme is validated by integrating it with state-of-the-art iterative attacks in multiple transfer scenarios. Experimental results corroborate that feature space fine-tuning can boost the transferability of existing targeted attacks nontrivially and universally. Our results also vote for the potential of simple iterative attacks to yield comparable targeted transferability to resource-intensive methods.

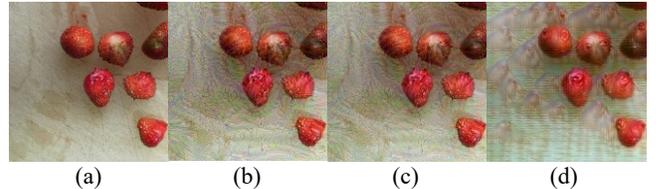

**Fig. 3.** The visual comparison of the AEs generated by different methods, $\epsilon = 16$. The target class is '*hippopotamus*'. (a) Original image, (b) Logit, (c) Logit+*ft* (ours), (d) TTP.

# Enhancing targeted transferability via feature space fine-tuning: supplementary material

*Hui Zeng, Biwei Chen, and Anjie Peng*

The supplementary document consists of five parts of content: A) Comparison with ILA [1], ILA++ [2, 3]; B) Ablation study on $N_{ft}$ and $k$; C) Visual comparison; D) Date-free targeted Universal adversarial perturbation (UAP); E) Alternative methods for calculating aggregate gradient.

*A Comparison with existing fine-tuning methods*

Several novel fine-tuning methods in untargeted attack have been proposed in the past few years. For instance, ILA maximizes the scalar projection of the adversarial example on the direction $f_l(\mathbf{I}') - f_l(\mathbf{I})$ , where $f_l()$ is the feature presentation in the $l$-th layer. ILA++ takes advantage of the directional guides gathered at each step of the baseline attack for a more stable guide direction. A straightforward question is:

*Is the proposed method merely a targeted version of existing fine-tuning methods [1, 2, and 3]?*

To answer this question, we first take a schematic diagram to illustrate the difference between the proposed method and the targeted version of ILA. As shown in Fig. 1, although both fine-tuning schemes start from an attacked image $\mathbf{I}'$ , the following directions are different. ILA is guided by $f_l(\mathbf{I}') - f_l(\mathbf{I})$, whereas the proposed method is guided by $f_l(\mathbf{I}')$. Such a difference embodies our unique considerations for targeted attacks:

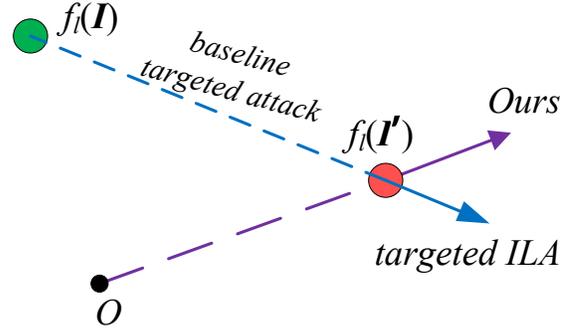

**Fig. 1.** Illustration of the different between the proposed method and targeted ILA. '$O$' represents the origin.

1) ILA is initially designed for untargeted attacks. Hence, it prioritizes moving away from the original class. In contrast, our strategy focuses more on increasing the probability of the target class.

2) We believe that the image content $\mathbf{I}$ and the adversarial perturbation $\mathbf{I}' - \mathbf{I}$ are closely intertwined in the feature space. Thus, $f_l(\mathbf{I}') - f_l(\mathbf{I})$ may not be a better guide than $f_l(\mathbf{I}')$.

Then, we experimentally compare the proposed method with targeted ILA. The experimental setup ($N_{ft}, k$) of the targeted ILA is the same as the proposed method. Tables 1 and 2 report the targeted success rates of compared schemes under the random-target and most difficult-target scenarios, respectively. The proposed method triumphs over targeted ILA by a clear margin.

**Table 1.** Comparison of fine-tuning with ILA and the proposed method. Targeted transfer success rates (%) in the single-model, random-target scenario. Dominant results are in **bold**.

| Attack | Source Model: Res50 | | | Source Model: Den121 | | | Source Model: VGG16 | | | Source Model: Inc-v3 | | | AVG |
|---|---|---|---|---|---|---|---|---|---|---|---|---|---|
| | →Inc-v3 | →Den121 | →VGG16 | →Inc-v3 | →Res50 | →VGG16 | →Inc-v3 | →Res50 | →Den121 | →Res50 | →Den121 | →VGG16 | |
| CE | 3.9 | 44.9 | 30.5 | 2.3 | 19.0 | 11.3 | 0.0 | 0.3 | 0.5 | 1.8 | 2.1 | 1.5 | 9.8 |
| CE+ILA | **10.6** | 60.3 | 44.2 | 11.5 | 35.6 | 29.3 | 0.0 | 1.6 | 3.2 | 2.6 | 4.3 | 2.8 | 17.2 |
| CE+*ft* (ours) | 9.0 | **60.4** | **49.3** | **13.2** | **45.3** | **34.8** | 0.0 | **2.8** | **3.3** | **4.7** | **7.8** | **4.0** | **19.6** |

**Table 2.** Comparison of fine-tuning with ILA and the proposed method. Targeted transfer success rates (%) in the single-model, most difficult-target scenario.

| Attack | Source Model: Res50 | | | Source Model: Den121 | | | Source Model: VGG16 | | | Source Model: Inc-v3 | | | AVG |
|---|---|---|---|---|---|---|---|---|---|---|---|---|---|
| | →Inc-v3 | →Den121 | →VGG16 | →Inc-v3 | →Res50 | →VGG16 | →Inc-v3 | →Res50 | →Den121 | →Res50 | →Den121 | →VGG16 | |
| CE | 1.3 | 25.8 | 15.0 | 1.2 | 6.5 | 3.6 | 0.0 | 0.0 | 0.0 | 2.4 | 4.2 | 2.3 | 5.2 |
| CE+ILA | **3.6** | 42.4 | 28.9 | 5.2 | 19.5 | 14.7 | 0.0 | 0.0 | 0.0 | 4.3 | 4.7 | **5.0** | 10.7 |
| CE+*ft* (ours) | 3.1 | **45.3** | **29.7** | **6.1** | **23.4** | **19.2** | 0.0 | **1.4** | **1.2** | **5.9** | **7.8** | **5.0** | **12.3** |

## B Ablation study

1) **Influence of the fine-tuning iteration** $N_{ft}$. We study the influence of $N_{ft}$ on the transfer success rate in the single-model, random-target scenario, with the source model fixed as Res50. The optimal $N_{ft}$ varies from 10 to 15 when the baseline attack is CE (Fig. 2(a)) and from 5 to 10 when the fine-tuning is based on Logit (Fig. 2(b)). This can be explained as follows. A relatively weak attack, e.g., CE, has greater potential for improvement and thus needs more iterations of fine-tuning. In contrast, a relatively strong attack, Logit or model-ensemble, is more suitable for less fine-tuning. In our study, we set $N_{ft} = 10$ for all attacks and in all scenarios for simplicity. Fig. 2 indicates that $N_{ft} = 10$ is almost always dominant $N_{ft} = 0$ that represents no fine-tuning.

2) **Influence of the fine-tuning layer $k$.** Next, we study the influence of target layer $k$ in fine-tuning on the transfer success rate. In this experiment, we fix the other parameters of the proposed method and select a few internal layers for each source model. Fig. 3(a), (b), and (c) report the transferability of adversarial examples crafted on source models Res50, Dense121, and Inc-V3, respectively. The main takeaway is that fine-tuning on a middle layer is helpful to transferability. This finding is consistent with previous works that early layers are usually data-specific, whereas later ones are model-specific. Based on the above considerations, we select to attack *Mixed_6b* for Inc-v3, the last layer of the third block for Res50 and Dense121, and *Conv4_3* for VGG16 in this study.

## B Visual comparison

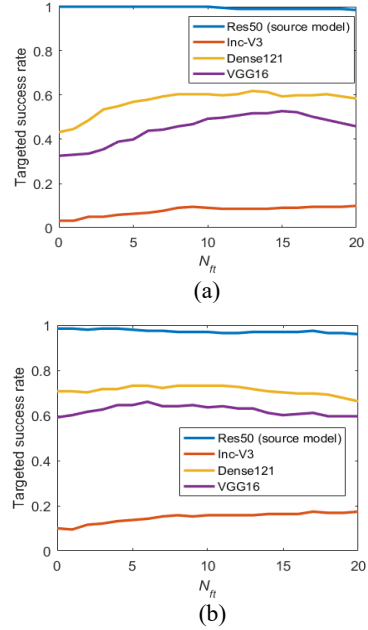

**Fig. 2.** Effect of $N_{ft}$ on AEs' transferability. The source model is Res50. The baseline attack is CE (a) and Logit (b).

Besides the example in the paper, we provide additional examples in this file. Fig. 4 shows AEs targeted to '*grey owl*,' and Fig. 5 shows AEs targeted to '*hippopotamus*.' While the perturbation introduced by the iterative methods resembles noise, that introduced by TTP is more semantically-aligned.

## C Date-free targeted UAP

Targeted UAP is a particular type of perturbation that can drive multiple clean images into a given class $y_t$. Among the methods for crafting targeted UAP, we are particularly interested in the data-free approach, which does not require additional training data. Precisely, we use a mean image (all entrances of which equal 0.5) as the start point and mount a targeted attack for 200 iterations to obtain a targeted UAP ($\epsilon = 16$) with different simple iterative methods. Then, the obtained UAP is applied to all 1000 images in our dataset.

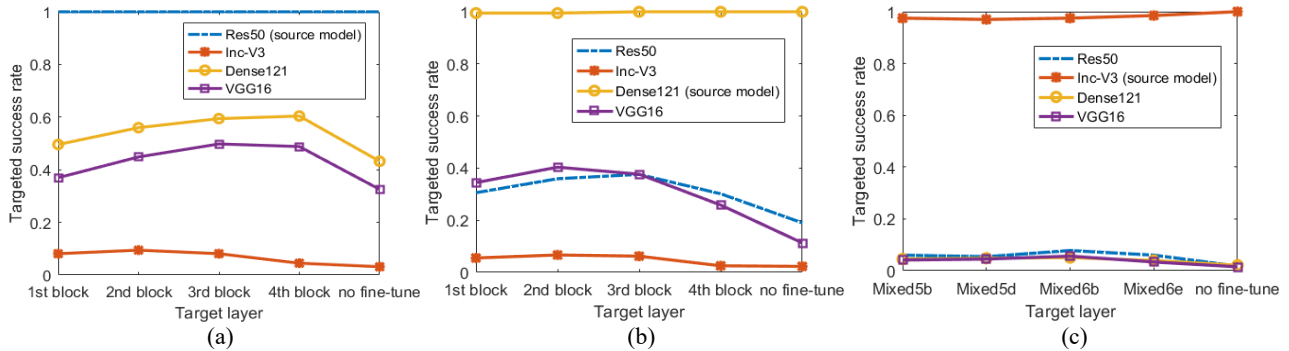

**Fig. 3.** Effect of target layer on AEs' transferability. The baseline attack is CE. The source models are Res50 (a), Dense121 (b), and Inc-V3 (c), respectively.

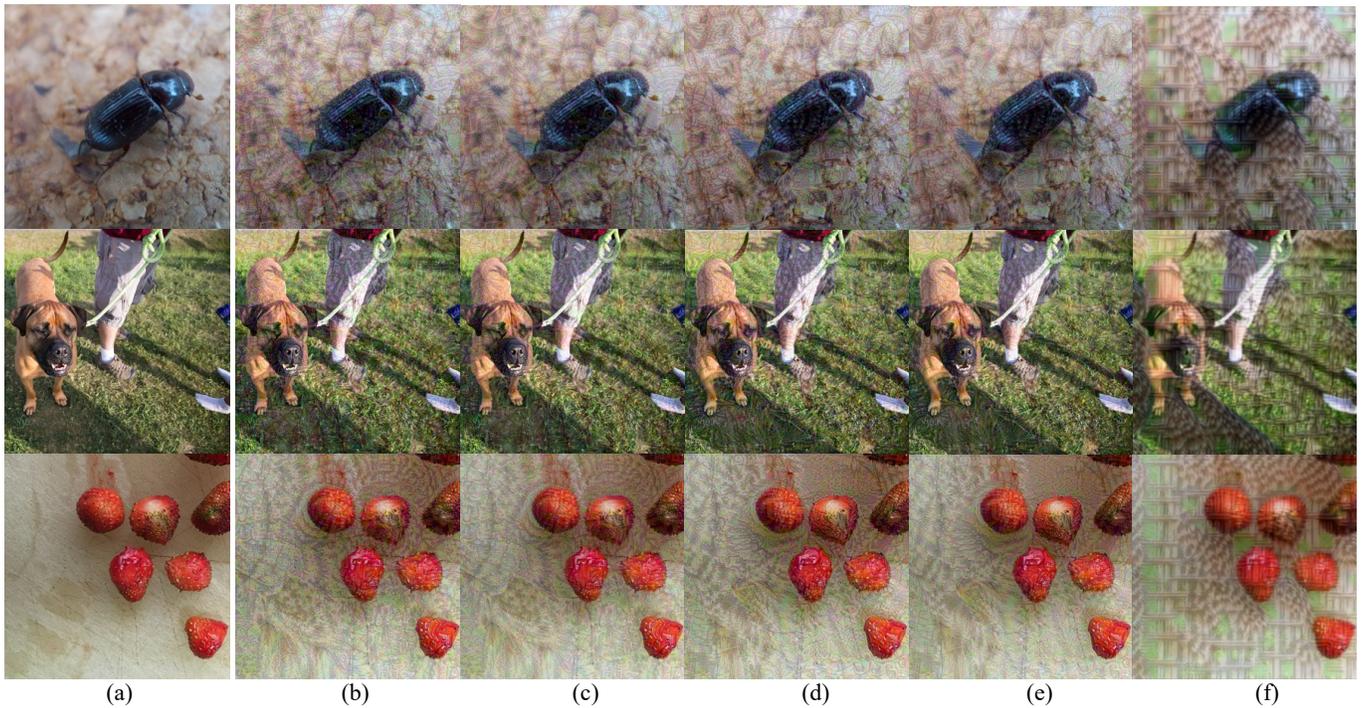

**Fig. 4.** The visual comparison of the AEs generated by different methods, $\epsilon = 16$. The target class is '*grey owl*'. (a) Original image, (b) CE, (c) CE+*ft* (proposed), (d) Po+Trip, (e) Po+Trip+*ft* (proposed), (f) TTP.

Table 3 reports the success rates averaged over 100 classes ($y_t = 0 : 99$). It is observed that feature space fine-tuning consistently improves the baseline attacks. Combining the results of the paper, we can conclude that the proposed fine-tuning scheme improves AEs' transferability not only across models but also across input images. Fig. 6 presents examples of targeted UAPs generated with different methods. It is observed that the UAPs are less noisy after feature space fine-tuning.

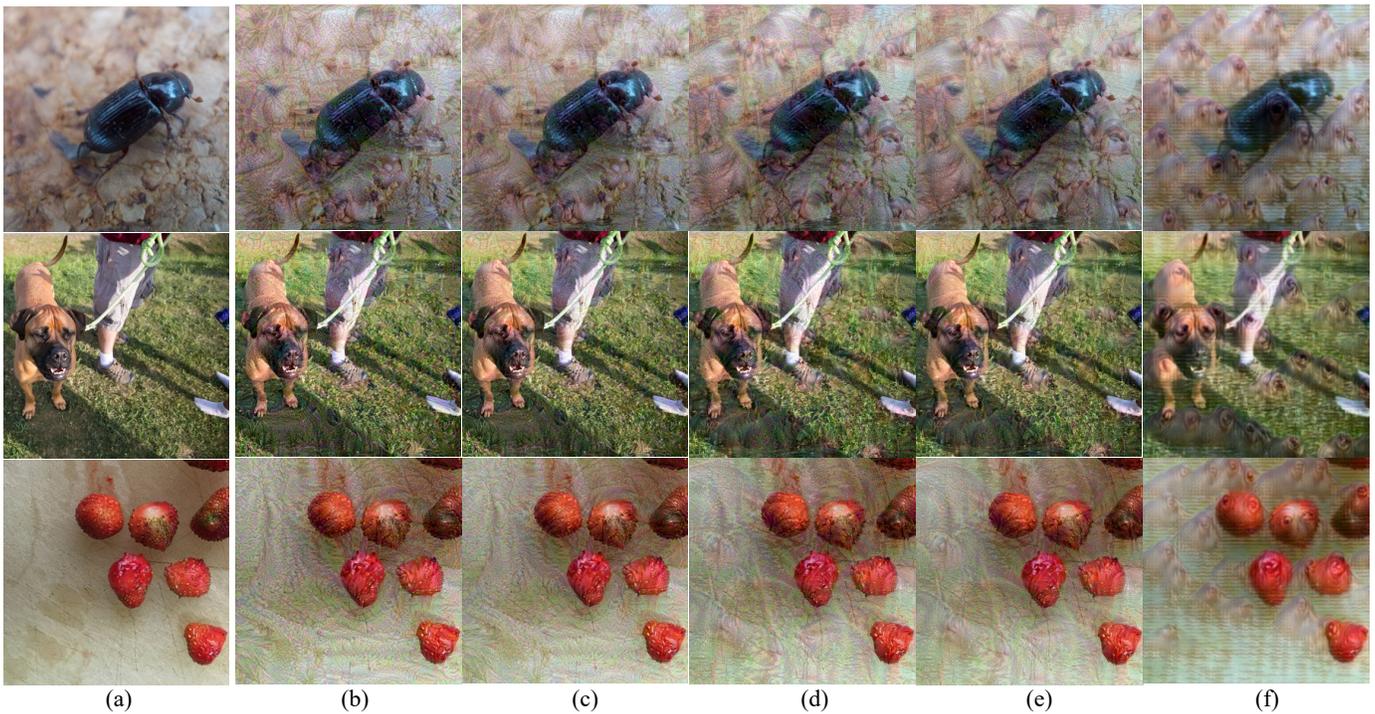

**Fig. 5.** The visual comparison of the AEs generated by different methods, $\epsilon = 16$. The target class is '*hippopotamus*'. (a) Original image, (b) Logit, (c) Logit+*ft* (proposed), (d) SupHigh, (e) SupHigh+*ft* (proposed), (f) TTP.

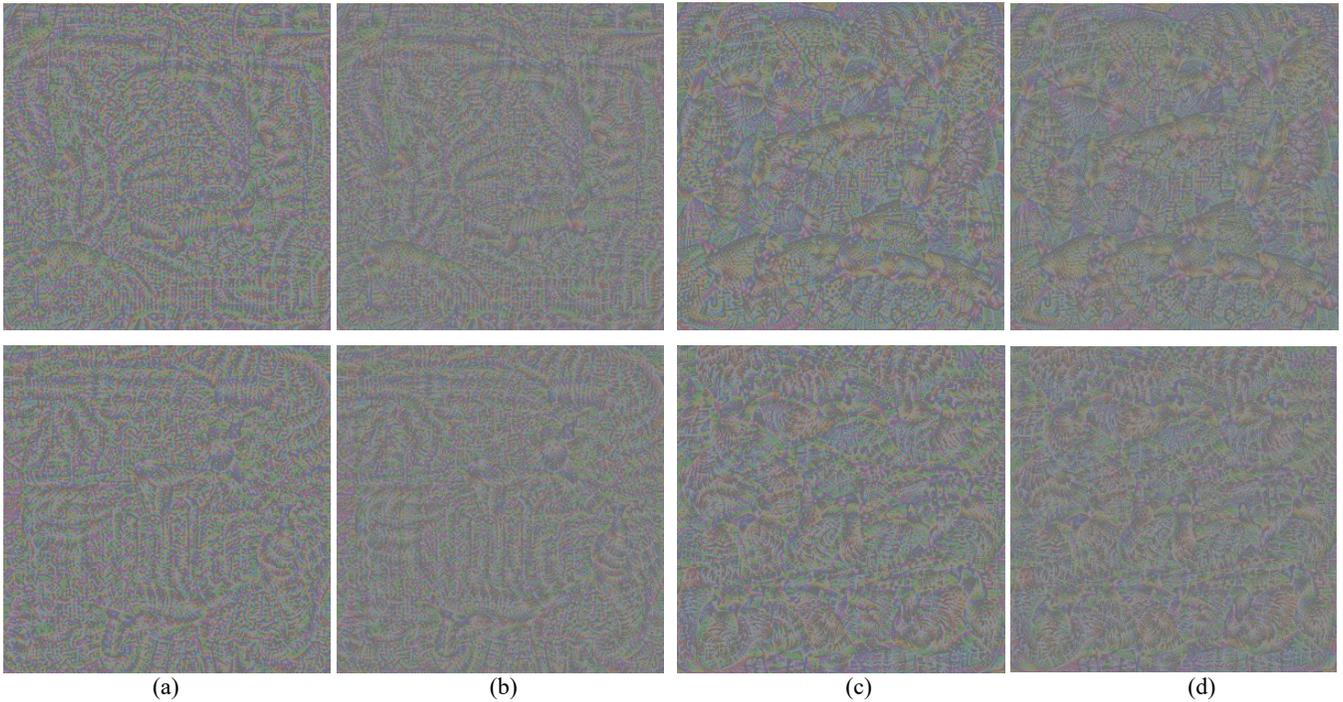

(a)         (b)         (c)         (d)

**Fig. 6.** Data-free targeted UAPs ($\epsilon = 16$,VGG16) generated by different methods. The target class is '*tench*' for the first row, and '*goose*' for the second row. (a) CE, (b) CE+*ft*, (c) Logit, (d) Logit+*ft* (proposed).

**Table 3.** Success rates (%) of the data-free UAPs with $\epsilon = 16$. Without/with fine-tuning. Dominant results are in **bold**.

| Attack | Res50 | Dense121 | VGG16 | Inc-v3 |
|--------|-------|----------|-------|--------|
| CE | 8.1/**15.1** | 8.0/**13.1** | 19.2/**34.6** | 1.9/**2.4** |
| Logit | 20.7/**24.6** | 17.5/**18.8** | 64.9/**66.3** | 3.6/**4.7** |

### D Alternative aggregate gradients

This subsection investigates the effect of the method of calculating aggregate gradients on the proposed fine-tuning scheme. Fig. 7 compares the transferability of AEs (under the random-target and most difficult-target scenarios, $\epsilon = 16$ ) when the aggregate gradient is generated with FIA [4] and RPA

[5]. Unlike FIA, which adopts a pixel-wise mask, RPA adopts a patch-wise mask in calculating aggregate gradients. For a fair comparison, we set the ensemble number $N$=30 for both FIA and RPA. Our results show that the more advanced RPA indeed improves the transferability slightly in most cases. This result indicates the proposed method can be further improved by incorporating more advanced aggregate gradient methods. In the paper, we use FIA to generate the aggregate gradient for simplicity.

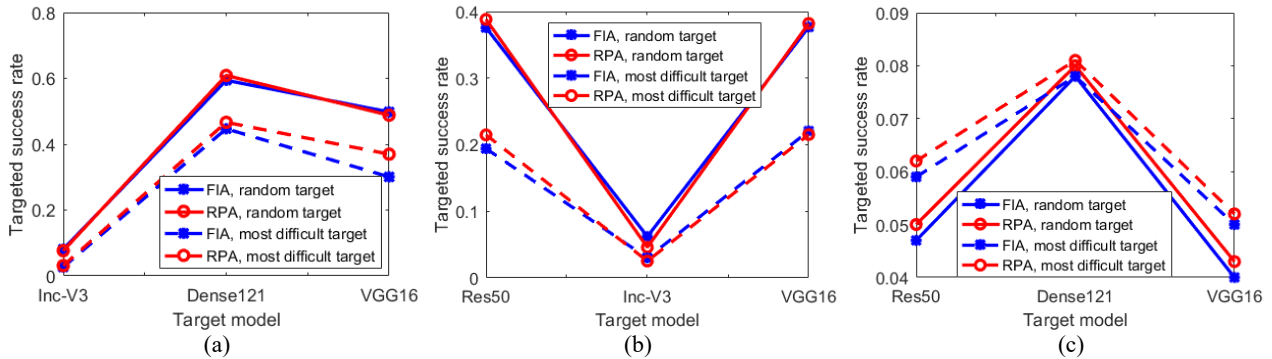

(a)        (b)        (c)

**Fig. 7.** Comparison AEs' transferability when the aggregate gradients are generated with FIA and RPA. The baseline attack is CE. The source models are Res50 (a), Dense121 (b) , and Inc-V3 (c), repectively.